\documentclass{Interspeech}

\interspeechcameraready

\title{Reasoning-Based Approach with Chain-of-Thought for Alzheimer’s Detection Using Speech and Large Language Models}

\author[affiliation={1}]{Chanwoo}{Park}
\author[affiliation={2}]{Anna Seo Gyeong}{Choi}
\author[affiliation={3}]{Sunghye}{Cho}
\author[affiliation={1\dagger}]{Chanwoo}{Kim}

\affiliation{Department of Artificial Intelligence}{Korea University}{South Korea}
\affiliation{Information Science}{Cornell University}{United States}
\affiliation{Department of Linguistics}{University of Pennsylvania}{United States}
\email{\{cksdn1290, chanwcom\}@korea.ac.kr}
\keywords{Chain-of-Thought, dementia detection, cue, speech recognition, large language model}

\usepackage{comment}
\usepackage{graphicx}
\usepackage{cite}
\usepackage{amsmath,amssymb,amsfonts}
\usepackage{algorithmic}
\usepackage{textcomp}
\usepackage{xcolor}
\usepackage{booktabs}
\usepackage{multirow}
\usepackage{svg}
\usepackage{mathtools}
\usepackage{subfig}
\usepackage{tabularx}
\usepackage{float}
\usepackage{placeins}
\usepackage{dblfloatfix}
\usepackage{pifont}

\begin{document}

\maketitle

\begin{abstract}
  Societies worldwide are rapidly entering a super-aged era, making elderly health a pressing concern.
  The aging population is increasing the burden on national economies and households.
  Dementia cases are rising significantly with this demographic shift.
  Recent research using voice-based models and large language models (LLM) offers new possibilities for dementia diagnosis and treatment.
  Our Chain-of-Thought (CoT) reasoning method combines speech and language models.
  The process starts with automatic speech recognition to convert speech to text. 
  We add a linear layer to an LLM for Alzheimer's disease (AD) and non-AD classification, using supervised fine-tuning (SFT) with CoT reasoning and cues.
  This approach showed an 16.7\% relative performance improvement compared to methods without CoT prompt reasoning.
  To the best of our knowledge, our proposed method achieved state-of-the-art performance in CoT approaches.
\end{abstract}

\renewcommand{\thefootnote}{\fnsymbol{footnote}}
\footnotetext[2]{Corresponding author}
\footnotetext[0]{The source code is available at GitHub by \texttt{https://doi.org/10.5281/zenodo.15511013}}
\renewcommand{\thefootnote}{\arabic{footnote}}

\section{Introduction}

Dementia is a neurodegenerative disorder characterized by the progressive decline of cognitive functions.
It significantly impacts patients' ability to perform daily activities, thereby severely affecting their quality of life.
According to the World Health Organization (WHO), approximately 50 million people worldwide are living with dementia, and nearly 10 million new cases are diagnosed each year \cite{livingston2020dementia}.
Early detection is crucial, as it can substantially reduce treatment costs and mitigate the disease's impact, improving outcomes for patients and caregivers alike.
Moreover, dementia not only poses challenges for individuals but also places a significant economic and emotional burden on families and healthcare systems globally.
Research suggests that raising awareness about dementia and improving access to diagnostic tools can enhance early intervention efforts.
Public health strategies focusing on prevention, such as promoting a healthy lifestyle, managing cardiovascular risk factors, and encouraging cognitive engagement, may also play a vital role in reducing the prevalence of dementia in aging populations.

According to recent research, efforts are being made to solve various medical problems using large language model (LLM).
In particular, various methods for the early diagnosis of Alzheimer's Disease (AD) using speech and verbal information have been proposed.
Researchers have identified acoustic and linguistic patterns experienced by AD patients, such as forgetting words, grammatical errors, increased pauses during speech, and repeated use of words.
Various studies are being conducted to classify patients using LLM.
For comparison with studies dealing with the same problem, the data provided in the Alzheimer's Dementia Recognition through Spontaneous Speech (ADReSS) dataset \cite{luz2021alzheimer} can be used to effectively classify AD.
The ADReSS dataset consists of voice recordings labeled as AD or non-AD, where patients describe a picture of a "cookie thief."

In this study, the pre-trained Whisper model can be used to transcribe speech into text and utilize linguistic information.
Fine-tuning is essential because state-of-the-art pre-trained LLMs show very strong performance but have limitations in highly specific tasks.
However, fine-tuning all parameters of an LLM poses challenges due to its large capacity. Updating all parameters can lead to the loss of existing knowledge within the LLM and requires significant computational resources and memory.

To address this issue, Parameter-Efficient Fine-Tuning (PEFT) \cite{ding2023parameter} is employed to update only a subset of parameters rather than all of them.
This approach helps preserve the existing knowledge of the LLM while reducing memory usage and computational costs. Among PEFT methods, this study selects Low-Rank Adaptation (LoRA) \cite{hu2021lora} fine-tuning.
LoRA works by learning additional low-rank parameters while keeping the existing weights of the model fixed.
In other words, instead of modifying the original weights directly, LoRA adds learnable parameters to capture new information.

Previous AD detection studies primarily relied on single-modality approaches using either acoustic or linguistic features, facing limitations in addressing both aspects simultaneously.
This study overcomes these constraints by employing a Chain-of-Thought (CoT) \cite{wei2022chain} methodology that strategically identifies and leverages critical cues within individual modalities, enhancing detection performance without relying on multimodal integration.

\begin{figure*}[htb!]
  \centering
  \includegraphics[width=1\linewidth]{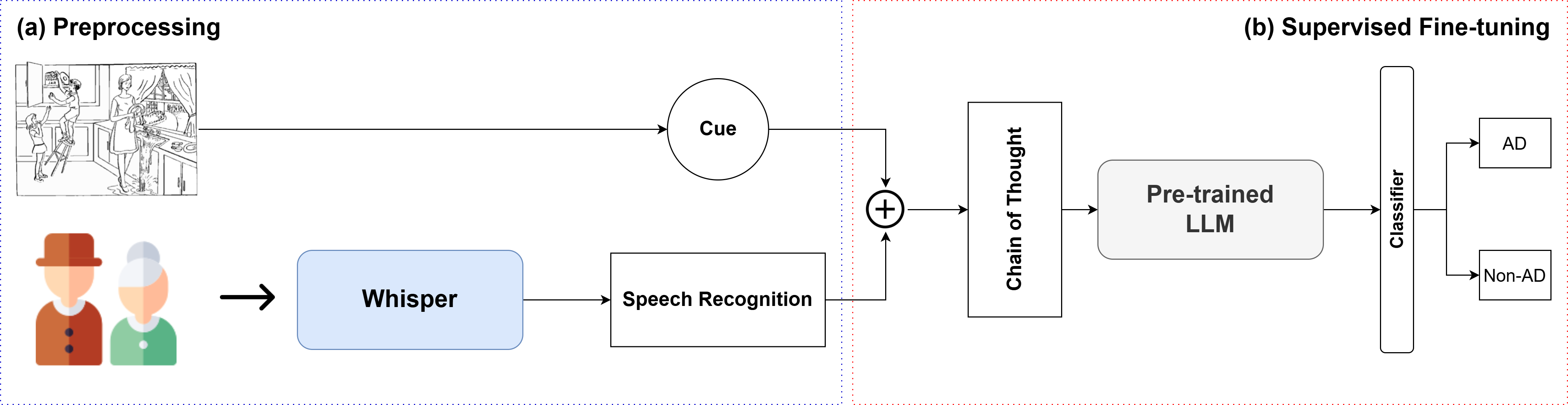}
  \caption{The process of LLM for dementia detection.
  (a) The preprocessing stage involves speech recognition of participant responses using a speech model and generation of important visual cues.
  (b) In the supervised fine-tuning (SFT) stage, utilize the pre-trained language model Llama to classify between AD and non-AD cases.}
  \label{fig:1}
\end{figure*}

\section{Related Works}
\label{sec:related}

Existing research on dementia diagnosis emphasizes the widespread use of cognitive screening tools like the mini-mental state examination (MMSE) and clinical dementia rating (CDR) in primary care settings, particularly in high-income countries \cite{pelegrini2019diagnosing}.
Diagnostic criteria from the DSM-IV/V and ICD-10 remain foundational, though middle-income countries often lack resources for advanced neuroimaging or biomarker testing \cite{pelegrini2019diagnosing}.
Recent advances highlight the potential of AI-driven retinal imaging to detect amyloid plaques for early diagnosis and autophagy-mediated tau protein degradation as a novel therapeutic target.
Biomarker frameworks like the ATN system (Amyloid-$\beta$, Tau, Neurodegeneration) are reshaping Alzheimer's disease diagnostics toward preclinical detection \cite{elahi2017clinicopathological, manap2024alzheimer}.
Studies consistently note significant underdiagnosis in primary care, exacerbated by inadequate clinician training and caregiver communication challenges \cite{pelegrini2019diagnosing, stokes2015dementia}.

Recent advancements in machine learning, speech processing, and natural language processing (NLP) have enabled the development of innovative approaches for detecting AD.
These methods leverage speech, language models, and multimodal data to identify early biomarkers of cognitive decline.
Below is a comprehensive summary of the two key approaches: speech-based methods and language model approaches.

\subsection{Speech-Based Approaches}

Speech analysis has emerged as a promising non-invasive tool for early detection of AD due to its strong correlation with cognitive health.
Researchers have identified several acoustic and linguistic features that differentiate individuals with AD from healthy controls.
Studies have shown that individuals with AD exhibit slower speech rates, longer pauses, increased hesitations, and more frequent grammatical errors compared to healthy individuals \cite{qi2023noninvasive, haider2019assessment}.
Feature sets like eGeMAPS (extended Geneva Minimalistic Acoustic Parameter Set) have proven effective in capturing these acoustic anomalies.
For instance, eGeMAPS \cite{chen2021automatic} achieved a classification accuracy of 74.1\% in distinguishing AD patients from healthy controls using leave-one-subject-out (LOSO) evaluation.
Advanced deep learning models such as Wav2Vec2 \cite{baevski2020wav2vec} and HuBERT \cite{hsu2021hubert} have further improved performance by learning robust speech representations from raw audio data \cite{chen2023exploring}.

\subsection{Language Model Approaches}

Models like Llama2-7B \cite{touvron2023llama} and Mistral-7B \cite{jiang2023mistral} have demonstrated impressive performance when fine-tuned for AD detection tasks \cite{casu2024optimizing}.
ChatGPT \cite{achiam2023gpt} has been used to analyze transcriptions of spontaneous speech, extracting linguistic patterns such as repetition, topic-switching, and fluency irregularities that are characteristic of AD \cite{bang2024alzheimer}.

\section{Methodology}

\subsection{Dataset}
Our experiment utilized data from the cookie theft picture description task of the Boston Diagnostic Aphasia Exam (BDAE) \cite{draper1973assessment, goodglass1983boston, goodglass2001bdae}, which is part of DementiaBank's Pitt Corpus \cite{luz2021alzheimer}.
The transcripts, annotated using the CHAT coding system \cite{macwhinney2009childes}, were acoustically enhanced through static noise removal.
Audio volume was normalized across all speech segments.
The dataset consists of speech samples from AD and non-AD English-speaking participants for the cookie theft picture description task, balanced by age, gender, and disease status, comprising a total of 156 participants with 1,955 speech segments from 78 non-AD participants and 2,122 speech segments from 78 AD participants.
The data was divided into a training set of 108 participants, consisting of 48 AD participants (55 minutes 46 seconds) and 48 non-AD participants (1 hour 14 minutes), and a test set of 48 participants (24 AD and 24 non-AD participants, 1 hour 6 minutes).
In speech processing, cookie theft detection (CTD) involves evaluating cognitive function by analyzing a person's ability to describe a specific image.

\subsection{Frameworks}

Our proposed dementia detection algorithm, which utilizes language and speech models, respectively, is illustrated in Figure~\ref{fig:1}.
First, we use Whisper \cite{radford2023robust}, a speech recognition model, to perform text transcription, converting spoken language into written text from participant's voice data.
Next, we generate relevant cues from the given cookie theft picture that could serve as important indicators.
Finally, we apply LoRA to a pre-trained LLM and perform supervised fine-tuning to classify whether an individual has AD or non-AD.

The data contains audio conversations between the investigator and participants.
Since the investigator tends to prompt responses when participants hesitate, we aimed to eliminate this influence.
As we wanted to focus solely on the participants' speech, we utilized the normalized audio chunks from each participant.
For speech recognition, we employed the large-v2 version of the Whisper model.

\subsection{Supervised Fine-tuning}

Supervised fine-tuning (SFT) is a crucial technique for adapting pre-trained LLMs to specific domains or tasks.
Using \texttt{Llama3.2-1B-Instruct} \cite{dubey2024llama}, as shown in Figure~\ref{fig:1}, the LLM takes text input to generate rich contextual embeddings, and a classification head, which is a linear layer, is added on top of the pre-trained LLM to distinguish between AD and non-AD cases.

\subsubsection{Chain-of-Thought}
Language models have shown various benefits when scaled up, including improved performance and sample efficiency.
However, scaling model size alone has proven insufficient to achieve high performance on challenging tasks like arithmetic, common sense, and symbolic reasoning.
Traditional few-shot prompting \cite{brown2020language} often proves ineffective for tasks requiring reasoning abilities and may not significantly improve even with larger language models.
Chain-of-thought (CoT) \cite{wei2022chain} explores language models' ability to perform few-shot prompting for reasoning tasks.
CoT consists of intermediate reasoning steps that lead to the final output.
This reasoning can be applied to tasks such as mathematical problems, common sense reasoning, and symbolic manipulation, and theoretically can potentially be applied to any task that humans can solve through language.
Since LLM generate sentences sequentially like filling in blanks, CoT prompting enables complex reasoning through intermediate reasoning steps.
When combined with direct answer prompting, it can yield better results in complex tasks that require reasoning before responding.

It then generates important cues from the CTD and integrates them into CoT prompts.
The cues entered at the CoT prompts are listed in Table~\ref{tab:cue}.
Specifically, we calculate the proportion of these listed cues expressed by participants in the text obtained from speech recognition, and provide prompts to infer whether the case is AD or non-AD.

\begin{table}[h!]
  \centering
  \begin{tabular}{ccc}
  \hline
  \multicolumn{3}{p{0.8\columnwidth}}{stool, sink, dish, wash, jar, cookie, child, mother, window, cabinet, kitchen, water} \\
  \hline
  \end{tabular}
  \caption{Cues for Chain-of-Thought (CoT) prompts.}
  \label{tab:cue}
\end{table}

\subsubsection{Parameter-Efficient Fine-Tuning}
Many natural language processing applications depend on adapting a single large pre-trained language model to various downstream applications.
One solution to this challenge is PEFT, with LoRA being a notable example.
LoRA is a specialized PEFT technique that efficiently fine-tunes LLM by adjusting only a small subset of parameters.
This method freezes the weights of the pre-trained model and inserts low-rank decomposition matrices into each layer, significantly reducing the number of trainable parameters.
This approach maintains performance comparable to full fine-tuning while decreasing computational and memory requirements.

In our experiments using PEFT, we configured LoRA with the following parameters: a rank value of 16 in the low-dimensional space, an adapter scaling value of 16, and a dropout probability of 0.01 prevent overfitting.

\begin{table}[htb!]
  \centering
  \begin{tabular}{l|c|c}
  \hline
  \textbf{Method} & \textbf{Acc (\%)} & \textbf{F1 (\%)} \\
  \hline
  Baseline & 75.00 & 74.83 \\
  Zero-shot & 47.92 & 32.39 \\
  Few-shot & 54.17 & 44.54 \\
  CoT & \textbf{83.33} & \textbf{83.22} \\
  \hline
  \end{tabular}
  \caption{Accuracy and F1-score results of our proposed method.}
  \label{tab:3}
\end{table}

\begin{table*}[htb!]
  \centering
  \begin{tabular}{cccccc}
  \hline
  \multicolumn{2}{c}{\multirow{2}{*}{\textbf{Model}}} & \multicolumn{2}{c}{\textbf{Pre-trained Model}} & \multirow{2}{*}{\textbf{Acc (\%)}} & \multirow{2}{*}{\textbf{F1 (\%)}} \\
  \cline{3-4}
  & & \multicolumn{1}{c}{\textbf{Speech}} & \multicolumn{1}{c}{\textbf{LM}} & & \\
  \hline
  \multicolumn{2}{c}{LDA \cite{luz2021alzheimer}} & \multicolumn{2}{c}{acoustic \& linguistic features} & 76.8 & 74.5 \\
  \multicolumn{2}{c}{SVM \cite{syed2021automated}} & \multicolumn{2}{c}{acoustic \& linguistic features} & 81.3 & - \\
  \multicolumn{2}{c}{Wav2Vec2 \cite{chen2023exploring}} & \checkmark & \ding{55} & 73.7 & 72.8 \\
  \multicolumn{2}{c}{HuBERT \cite{chen2023exploring}} & \checkmark & \ding{55} & 74.2 & 73.6 \\
  \multicolumn{2}{c}{BERT \cite{balagopalan2021comparing}} & \ding{55} & \checkmark & 83.3 & 83.9 \\
  \multicolumn{2}{c}{Bi-LSTM \cite{cummins2020comparison}} & \ding{55} & \checkmark & 81.3 & 81.2 \\
  \multicolumn{2}{c}{BERT + SpeechBERT \cite{zhu2021exploring}} & \checkmark & \checkmark & 82.9 $\pm$ 1.6 & 82.9 $\pm$ 1.9 \\
  \hline
  \multicolumn{2}{c}{\multirow{2}{*}{Ours (\texttt{Llama3.2-1B})}} & \checkmark & \checkmark & 83.3 & 83.2 \\
  \multicolumn{2}{c}{} & \ding{55} & \checkmark & \textbf{87.5} & \textbf{87.5} \\
  \hline
  \end{tabular}
  \caption{Experimental performance comparison using models of different modalities on the ADReSS Challenge set.}
  \label{tab:results}
\end{table*}

\section{Experiment results and analysis}
\label{sec:setup}

\subsection{Experimental Setup}

For our experiments, we trained the model using the original split of 108 participants for the SFT training set, and conducted evaluations using the remaining 48 participants in the test set.
We trained the model on a NVIDIA GeForce RTX 4090 GPU.
For the hyperparameter configuration, we used a batch size of 8 and fine-tuned the model with a learning rate of 1e-4 over three phases using the AdamW optimizer \cite{loshchilov2017decoupled}.
We applied a linear learning rate scheduler that incorporates a sustain phase—keeping the learning rate constant after the warm-up period before it decays linearly—and set the weight decay to 0.001.

\subsection{Results}

\begin{figure}[htb!]
  \centering
  \includegraphics[width=1\linewidth]{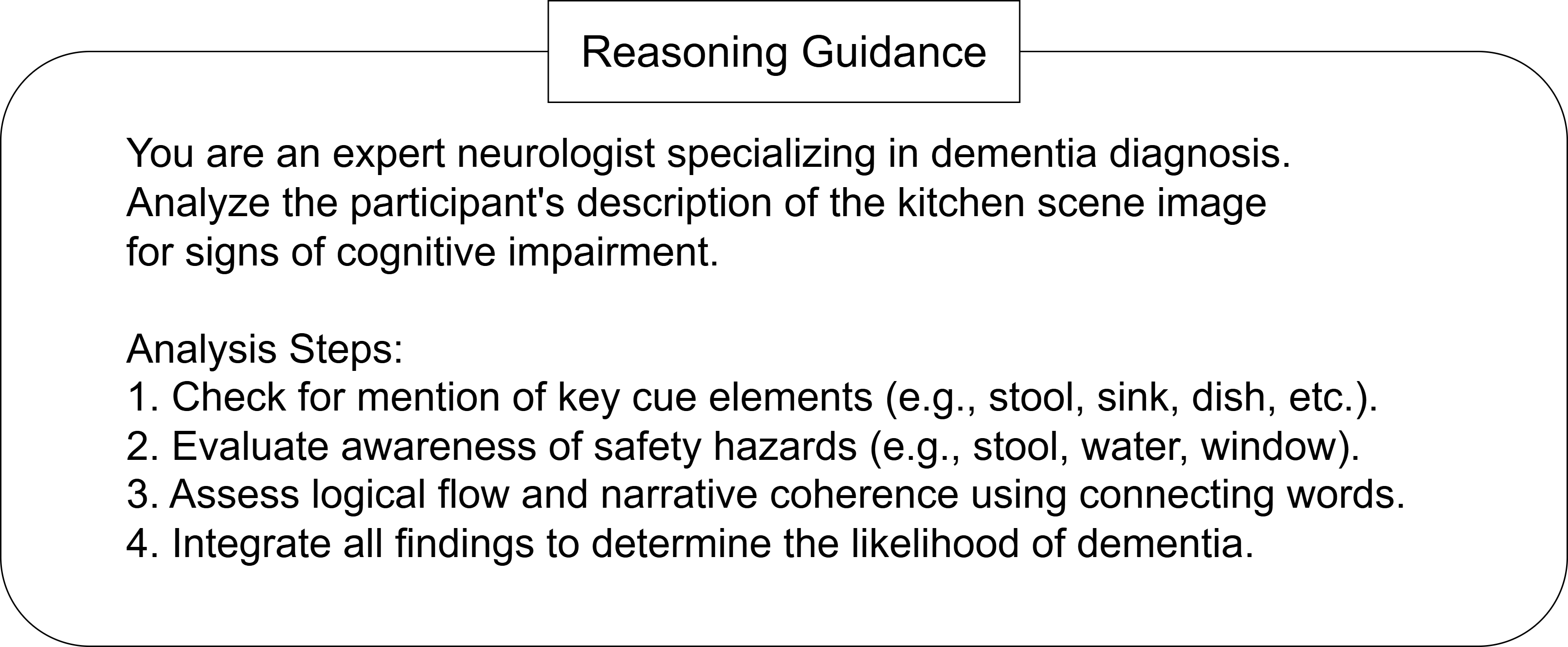}
  \caption{Example of the reasoning guidance Chain-of-Thought (CoT) prompts provided to the model for diagnosing dementia.}
  \label{fig:2}
\end{figure}

We conducted experiments comparing the baseline model trained with SFT without CoT prompts against zero-shot, few-shot, and our proposed CoT approaches.
The CoT prompts we used are shown in Figure~\ref{fig:2}.
The experimental results of the proposed LLM model are shown in Table~\ref{tab:3}.
The baseline results showed comparable performance to previous studies \cite{luz2021alzheimer}, while both zero-shot and few-shot settings demonstrated poor performance.
In contrast, we observed superior performance when incorporating CoT prompts that guide the reasoning process.  

\subsection{Comparison of results with other studies}

The Table~\ref{tab:results} compares the performance of various models across different modalities and datasets for the ADReSS \cite{luz2021alzheimer} benchmark.
Traditional methods \cite{luz2021alzheimer, syed2021automated} like linear discriminant analysis (LDA) \cite{xanthopoulos2013linear}, which rely on acoustic and linguistic features, achieve accuracy and F1-scores of 76.8\% and 74.5\%, respectively, indicating moderate performance \cite{luz2021alzheimer}.
This study \cite{syed2021automated}, utilizing a multimodal system, identified linguistic and paralinguistic characteristics of dementia through automated screening tools.
The researchers achieved an accuracy of 81.3\% using support vector machines (SVM) \cite{hearst1998support}, demonstrating that deep neural network embeddings and ensemble learning are viable approaches for objective dementia assessment.

This study \cite{chen2023exploring} utilizing spontaneous speech in AD detection demonstrates a non-invasive and cost-effective approach by implementing self-supervised learning (SSL) \cite{liu2021self} models with joint fine-tuning strategies, combined with multitask learning and data augmentation.
Fine-tuning pre-trained SSL models, along with multitask learning and data augmentation, enhances the effectiveness of universal speech representations in AD detection.
Speech-based models \cite{chen2023exploring} such as Wav2Vec2 \cite{baevski2020wav2vec} and HuBERT \cite{hsu2021hubert} show slightly lower results, with Wav2Vec2 achieving 73.7\% accuracy and 72.8\% F1-score, while HuBERT scores 74.2\% accuracy and 73.6\% F1-score, suggesting limited improvement over traditional approaches.

The study \cite{balagopalan2021comparing}, which utilized the effectiveness of speech transcript representations obtained from the BERT natural language processing model, compared it with more clinically interpretable language feature-based methods.
Both the feature-based approach and fine-tuned BERT model achieved an accuracy of 83.3\% and an F1-score of 83.9\% using a limited number of linguistic features.

Subsequently, in a text-based study \cite{zhou2016attention}, they experimented with a hierarchical neural network equipped with an attention mechanism trained on linguistic features, along with an acoustic-based system.
Using bidirectional long short-term memory (bi-LSTM) with attention \cite{cummins2020comparison}, they achieved performance scores of 81.3\% and 81.2\% for accuracy and F1-score, respectively.

A study \cite{zhu2021exploring} using dual-modality BERT and SpeechBERT \cite{chuang2019speechbert} models explored transfer learning techniques for AD classification and MMSE regression tasks.
The transfer learning models were pre-trained on general large-scale datasets and fine-tuned and tested using the ADReSS dataset, achieving 82.9 $\pm$ 1.56 and 82.9 $\pm$ 1.86 for accuracy and F1-score, respectively.

Finally, the proposed model using the ground truth (GT) text format recorded by the CHAT coding system and fine-tuned with CoT prompts achieved competitive results with an accuracy of 87.50\% and an F1-score of 87.48\%.
Our approach demonstrates robust performance across various input types, highlighting its versatility in handling complex data integration tasks.
Additionally, by employing the lightweight \texttt{Llama3.2-1B} model with fewer parameters, the approach ensures efficiency without compromising performance.

\subsection{Ablation Study}

The results of ablation study are presented in Table~\ref{tab:6}.
We fine-tuned the model using cues and CoT prompts, then compared the speech-to-text transcriptions with the ground truth content recorded using the CHAT coding system.
While the performance of the ASR model is crucial, considering the challenges in recognizing speech from middle-aged and elderly participants, we found that fine-tuning the LLM with the transcribed text yielded positive results.
  
\begin{table}[h!]
  \centering
  \begin{tabular}{ccc}
  \hline
  \textbf{Method} & \textbf{Acc (\%)} & \textbf{F1 (\%)} \\
  \hline
  Baseline & \multirow{2}{*}{83.33} & \multirow{2}{*}{83.22} \\
  (ASR CoT) & & \\
  \hline
  ground truth (SFT) & 83.33 & 83.30 \\
  ground truth (CoT) & \textbf{87.50} & \textbf{87.48} \\
  \hline
  \end{tabular}
  \caption{Comparison of accuracy and F1-score for different training strategies in the ablation study.
  The Baseline (ASR CoT) uses automatic speech recognition outputs without ground truth supervision.
  The ground truth supervised fine-tuning (SFT) is performed without Chain-of-Thought (CoT) prompts, while ground truth (CoT) applies CoT prompts during supervised fine-tuning.}
  \label{tab:6}
\end{table}

\section{Conclusions}
\label{sec:conclusion}

Advancements in AI technology, particularly LLM and multimodal approaches, are driving transformative changes in dementia research and treatment.
Models that combine automatic speech recognition (ASR) have demonstrated a 11.1\% improvement in accuracy for early diagnosis and classification of Alzheimer’s disease (AD) compared to traditional methods, underscoring the value of multimodal approaches.
The Chain-of-Thought (CoT) \cite{wei2022chain} reasoning framework enhances AI performance and transparency by breaking down complex diagnostic tasks into structured steps, thereby increasing its clinical applicability.
Additionally, speech-based AI models offer a non-invasive and scalable method for monitoring cognitive decline \cite{bang2024alzheimer}, making them a valuable tool for early detection and management in aging societies.
AI frameworks that integrate cues and neuroimaging data not only achieve high diagnostic accuracy but also contribute to predicting disease progression.

\section{Acknowledgment}
This work was supported by the IITP(Institute of Information \& Coummunications Technology Planning \& Evaluation)-ITRC(Information Technology Research Center)
grant funded by the Korea government(Ministry of Science and ICT)(IITP-2025-RS-2024-00436857),
IITP grant funded by the Korea government(MSIT) (No. RS-2019-II190079, Artificial Intelligence Graduate School Program(Korea University)),
the "Leaders in INdustry-university Cooperation 3.0" Project, supported by the Ministry or Education and National Research Foundation of Korea,
and IITP under the artificial intelligence star fellowship support program to nurture the best talents (IITP-2025-RS-2025-02304828) grant funded by the Korea government(MSIT).

\bibliographystyle{IEEEtran}
\bibliography{mybib}

\end{document}